\documentclass[conference]{IEEEtran}
\usepackage{times}
\usepackage[numbers]{natbib}
\usepackage{multicol}
\usepackage[bookmarks=true]{hyperref}
\hypersetup{
    urlcolor=blue,
}

\usepackage[utf8]{inputenc}
\usepackage{graphicx}
\usepackage{amsmath}
\usepackage{amssymb}
\usepackage{latexsym}
\usepackage{epsfig}
\usepackage{url}
\usepackage{booktabs}
\usepackage{multirow}
\usepackage{wrapfig}
\usepackage{xcolor}
\usepackage{todonotes}
\usepackage{subfigure}

\newif\ifpaperfinal
\paperfinalfalse
\ifpaperfinal
\newcommand{\tnnote}[1]{}
\newcommand{\stnote}[1]{}
\newcommand{\ellie}[1]{}
\newcommand{\ngnote}[1]{}
\else
\newcommand{\tnnote}[1]{\textcolor{teal}{\textbf{TN: #1}}}
\newcommand{\stnote}[1]{\textcolor{blue}{\textbf{ST: #1}}}
\newcommand{\ellie}[1]{\textcolor{magenta}{\textbf{EP: #1}}}
\newcommand{\ngnote}[1]{\textcolor{cyan}{\textbf{NG: #1}}}

\fi

\begin{document}

\title{Robot Object Retrieval\\with Contextual Natural Language Queries}

\author{Thao Nguyen, Nakul Gopalan, Roma Patel, Matt Corsaro, Ellie Pavlick, Stefanie Tellex\\
\{thaonguyen, romapatel, matthew\_corsaro, ellie\_pavlick\}@brown.edu\\
nakul\_gopalan@gatech.edu, stefie10@cs.brown.edu}

\date{}
\maketitle
\IEEEpeerreviewmaketitle

\begin{abstract}
Natural language object retrieval is a highly useful yet challenging task for robots in human-centric environments. 
Previous work has primarily focused on commands specifying the desired object's type such as ``scissors" and/or visual attributes such as ``red," thus limiting the robot to only known object classes. 
We develop a model to retrieve objects based on descriptions of their usage. 
The model takes in a language command containing a verb, for example ``Hand me something to \textit{cut}," and RGB images of candidate objects and selects the object that best satisfies the task specified by the verb.
Our model directly predicts an object's appearance from the object's use specified by a verb phrase. We do not need to explicitly specify an object's class label. Our approach allows us to predict high level concepts like an object's utility based on the language query.
Based on contextual information present in the language commands, our model can generalize to unseen object classes and unknown nouns in the commands. Our model correctly selects objects out of sets of five candidates to fulfill natural language commands, and achieves an average accuracy of 62.3\% on a held-out test set of unseen ImageNet object classes
and 53.0\% on unseen object classes \textit{and} unknown nouns. Our model also achieves an average accuracy of 54.7\% on unseen YCB object classes, which have a different image distribution from ImageNet objects.
We demonstrate our model on a KUKA LBR iiwa robot arm, enabling the robot to retrieve objects based on natural language descriptions of their usage\footnote{Video recordings of the robot demonstrations can be found at \url{https://youtu.be/WMAdGhMmXEQ}.}. We also present a new dataset of 655 verb-object pairs denoting object usage over 50 verbs and 216 object classes\footnote{The dataset and code for the project can be found at \url{https://github.com/Thaonguyen3095/affordance-language}.}.

\end{abstract}


\section{Introduction}
A key bottleneck in 
widespread deployment of robots in human-centric environments is the ability for non-expert users to communicate with robots. Natural language is one of the most popular communication modalities due to the familiarity and comfort it affords a majority of users. However, training a robot to understand open-ended natural language commands is challenging since humans will inevitably produce words that were never seen in the robot's training data. These unknown words can come from paraphrasing such as using ``saucer'' instead of ``plate,'' or from novel object classes in the robot's environments, for example a kitchen with a ``rolling pin'' when the robot has never seen a rolling pin before.


\begin{figure}[htbp]
    \centering
    \includegraphics[width=1.0\linewidth]{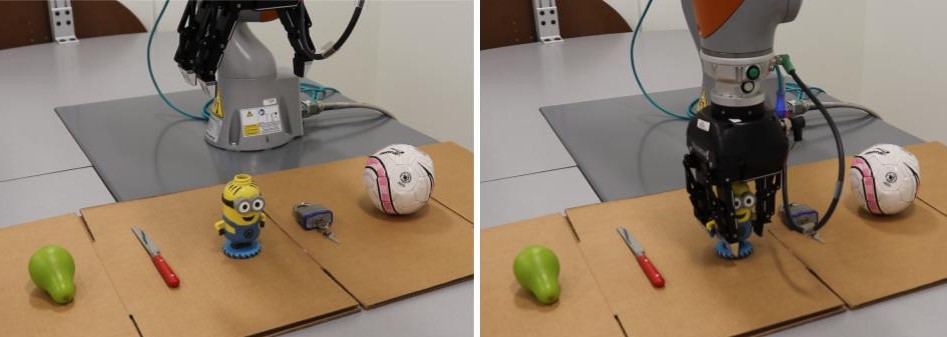}
    \caption{Our robot receives segmented RGB images of the objects in the scene and a natural language command such as ``Give me something to contain," and correctly retrieved the Minion (yellow cartoon character)-shaped container.}
    \label{fig:robot1}
\end{figure}

We aim to develop a model that can handle open-ended commands with unknown words and object classes. As a first step in solving this challenging problem, we focus on the natural language object retrieval task --- selecting the correct object based on an indirect
natural language command with constraints on the functionality of the object.
More specifically, our work focuses on fulfilling commands requesting an object for a task specified by a verb such as ``Hand me a box cutter to \textbf{cut}." 
Being able to handle these types of commands is highly useful for a robot agent in human-centric environments, as people usually ask for an object with a specific usage in mind.
The robot would be able to correctly fetch the desired object for the given task, such as cut,
without needing to have seen the object, a box cutter for example,
or the word representing the object, such as the noun ``box cutter."
In addition, the robot has the freedom to substitute objects as long as the selected object satisfies the specified task. This is particularly useful in cases where the robot cannot locate the specific object the human asked for but found another object that can satisfy the given task, such as a knife instead of a box cutter to cut.

There has been much prior work on natural language object retrieval \cite{krishnamurthy2013jointly, hu2016natural, chen2018text2shape, cohen2019grounding} and similar areas such as image captioning and image retrieval \cite{patterson2012sun, mao2014deep, vinyals2015show, xu2015show}.
However, previous work primarily focuses on natural language commands that either specify the object class such as ``scissors" or describe the object’s visual attributes such as ``red," ``curved," ``has handle,” and cannot handle unknown object classes or words. Our work, in contrast, anchors the desired object to its usage (specified by a verb) and reasons about the verb to handle unknown objects and nouns on-the-fly.
Our work demonstrates that an object's appearance provides sufficient signals to predict whether the object is suitable for a specific task, without needing to explicitly classify the object class and visual attributes. Our model takes in RGB images of objects and a natural language command containing a verb, generates embeddings of the input language command and images, and selects the image most similar to the given command in embedding space. The selected image should represent the object that best satisfies the task specified by the verb in the command. We train our model on natural language command-RGB image pairs. The evaluation task for the model is to retrieve the correct object from a set of five images, given a natural language command. We use ILSVRC2012 \cite{ILSVRC15} images and language commands generated from verb-object pairs extracted from Wikipedia for training and evaluation of our model. Our model achieves an average retrieval accuracy of 62.3\% on a held-out test set of unseen ILSVRC2012 object classes and 53.0\% on unseen object classes \textit{and} unknown nouns. Our model also achieves an average accuracy of 54.7\% on unseen YCB object classes. We also demonstrate our model on a KUKA LBR iiwa robot arm, enabling the robot to retrieve objects based on natural language commands, and present a new dataset of 655 verb-object pairs denoting object usage over 50 verbs and 216 object classes.

\section{Related Work}
Natural language object retrieval refers to the task of finding and recovering an object specified by a human user using natural language. The computer vision and natural language grounding communities attempt to solve object retrieval by locating or \textit{grounding} the object specified in an image using natural language \cite{krishnamurthy2013jointly, hu2016natural}. 
\citet{krishnamurthy2013jointly} use a dataset of RGB images with segmented objects and their natural language descriptions to learn the grounding of words to objects in the image by exploiting repeated occurrences of segmented objects within images, along with their descriptions in natural language. \citet{hu2016natural} use a similar approach albeit using deep neural networks to avoid parsing and feature construction by hand. \citet{chen2018text2shape} learn joint embeddings of language descriptions and colored 3D objects for text-to-shape retrieval and generation of colored 3D shapes from natural language. \citet{cohen2019grounding} learn joint embeddings of language descriptions and segmented depth images of objects for object retrieval within instances of the same object class. Our work, in contrast, learns an embedding across object classes based on their suitability for a given task specified using natural language. Our object embeddings are not conditioned on the output class of objects, but on the relevancy of the object for the specified task. This, therefore, allows us to retrieve objects based on descriptions of their usage and importantly allows handling of unknown nouns and unseen object classes.

Another relevant line of work is image captioning and image retrieval, which also aims to jointly model a natural language sequence and image content. The SUN scene attribute dataset \cite{patterson2012sun} maps images to attributes such as ``hills," ``houses," ``bicycle racks," ``sun," etc. Such understanding of image attributes provides scene category predictions and high level scene descriptions, for example ``human hiking in a rainy field." Methods based on recurrent neural networks (RNNs) \cite{mao2014deep, vinyals2015show, xu2015show} trained to directly model the probability distribution of generating a word given previous words and an image have shown to be effective in image caption generation, natural language image retrieval, and image caption retrieval. Our work is most similar to earlier attribute based image retrieval work. However, these models are trained on attributes that are directly specified and not inferred from indirect task based queries. Implicit task-based object attributes are harder to learn but are also more general than directly specifiable object visual attributes, and are useful for a natural language object retrieval system to have in its toolbox.

Also related to our work are methods on interactive object retrieval \cite{whitney2017reducing} and language grounding \cite{shridhar2018interactive, hatori2018interactively}. These methods perform inference over dialogue to deduce the right object based on the specifications provided by the human user. They specifically use known object copra and directly specify the object attributes. Our work, in contrast, retrieve objects based on contextual information about the task being specified by the natural language command. We are not performing inference over dialogue, but it is a natural next step for our work where our joint embedding can prove useful in the case of novel objects.


Similar to previous work,
our work aims to learn joint object representations from visual and language information using RNNs. However,  previous work primarily focuses on natural language commands specifying the object type and visual attributes, such as ``scissors," ``red," ``curved," ``has handle." In contrast, our work focuses on fulfilling commands requesting an object for a task specified by a verb, for example ``Hand me something to \textbf{cut}." 
To handle such commands, a possible approach is to rely on accurate classification of the object type and visual attributes and an external knowledge base to query for valid verb-object or verb-attribute pairings. However, that approach would be limited to known objects and words.
Our work, on the other hand, bypasses explicit classification of object type and attributes, and directly maps object use that is specified by the verb to object appearance that is captured by the image. Our work uses the context of the verb to implicitly infer object attributes that are required for the task, and can generalize to unseen object classes and unknown nouns in the language commands.

\section{Approach}

To fulfill natural language commands requesting an object for a task specified by a verb, our model generates embeddings for the language command and candidate objects and selects the object that is closest to the command in embedding space. Our model is trained using pairs of natural language object requests containing verbs and ground truth objects that satisfy the requests. We describe our model and data collection process in detail in Sections \ref{sec:model} and \ref{sec:data_collection}.

\subsection{Model}
\label{sec:model}

\begin{figure}[t]
\begin{center}
\includegraphics[width=1.0\linewidth]{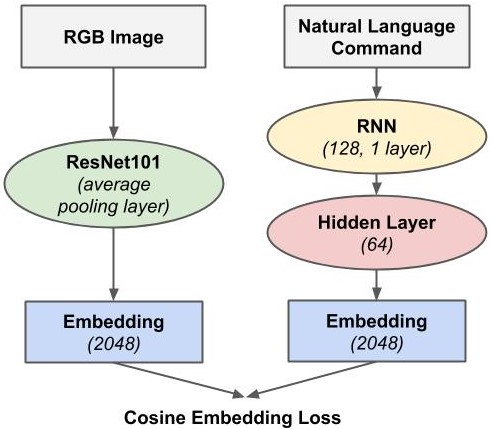}
\end{center}
   \caption{Diagram of the language-vision embedding model. The model encodes given natural language commands and RGB images, and minimizes the cosine embedding loss between the language and image embeddings during training. At inference time, the model calculates the cosine similarities between the embeddings of the language command and candidate images, and selects the image most similar to the command in embedding space.}
\label{fig:model}
\end{figure}

Given a natural language command and images of candidate objects, we want our model to correctly select the object that best satisfies the command. Our model does this by generating embeddings for the input natural language command and images, calculating the cosine similarities between the image embeddings and the language embedding, and selecting the image most similar to the command in embedding space. A diagram of our model is shown in Figure \ref{fig:model}. Our model consists of separate image and language encoders, each in charge of generating embeddings for the input natural language commands and RGB images, respectively. During training, our model minimizes the cosine embedding loss between the embeddings of language command-RGB image pairs, thus maximizing the likelihood of the target image given the command. We describe the component image and language encoders and our model training process below.

\subsubsection{Image Encoder}
To encode each RGB image, we use the average pooling layer of a pretrained ResNet101 \cite{he2016deep}. We chose ResNet101 due to ResNet's good performance on robots \cite{mallick2018deep}. The use of deep pretrained representations enables our model to leverage prior information of complex image features to allow for better encoding of the visual information from the images. We, therefore, get an embedding of size 2048 from the pretrained ResNet model for each RGB image.

\subsubsection{Language Encoder}
To encode each natural language command, our language encoder consists of a recurrent neural network (RNN) \cite{rnn} followed by a fully connected layer. We randomly initialize word embeddings for each language command that are then trained from scratch. The model, therefore, produces an embedding vector that is the same size of the embedding produced by the image encoder.

\subsubsection{Training Process}
We train our model to optimize an objective function that attempts to bring the corresponding language and image embeddings closer to each other in embedding space. We achieve this by reducing the cosine embedding loss between the low-dimensional embeddings produced by the image encoder, which takes in an RGB image of the object, and the embedding produced by the language encoder, which takes in the referring natural language command, during training.

We describe the training data in Section \ref{sec:data}. Positive training samples consist of pairs of natural language commands, each containing one verb, and RGB images of objects that can be paired with that verb. We obtain negative samples by randomly sampling an image of a different object that does not correspond with the verb and pairing the image with the language command, resulting in a dataset of equally balanced positive and negative samples. We use Adam \cite{kingma2014adam} as an optimiser with a learning rate of 0.0001 and train for 50 epochs until convergence.

\subsection{Data Collection}
\label{sec:data_collection}
To train and evaluate our model, we require pairs of natural language commands containing verbs and RGB images of objects. To obtain these command-image pairs, we need verb-object pairs denoting valid object usage such as ``cut" for a ``knife." We also require RGB images for the objects. Since we are interested in testing our model's generalization capability on unseen object classes and nouns, we require a large number of object classes that can be paired with the verbs for a sufficient number of held-out object classes.

To the best of our knowledge, no existing datasets of verb-object pairs met our requirements. \citet{chao2015mining} mine the web for the knowledge of semantic affordance --- given an object, determining whether an action can be performed on it --- resulting in a dataset of verb-noun combinations. However, their dataset is only on 20 object classes, and focuses on verbs denoting actions that can be performed on the objects, such as ``hunt" a ``bird," rather than the objects' usage. Other works on semantic affordances \cite{myers2015affordance, do2018affordancenet} also provide datasets of objects labeled with their affordances. However, these datasets are on fewer than 20 object classes and fewer than 10 affordances.

We, therefore, decided to collect our own dataset of valid verb-object pairs and use it to generate natural language commands paired with RGB images for our model. We describe our data below.

\begin{table}[]
\begin{center}
\caption{Example verb-object pairs from our dataset}
\begin{tabular}{llll} 
\toprule
contain -- bucket & hit -- hammer & wear -- necklace\\
contain -- wardrobe & hit -- racket & wear -- suit\\
cut -- cleaver & play -- baseball & wrap -- cloak\\
cut -- hatchet & play -- violin & wrap -- handkerchief\\
eat -- banana & serve -- plate & write -- notebook\\
eat -- pizza & serve -- tray & write -- quill\\
\bottomrule
\label{tab:verb-obj}
\end{tabular}
\end{center}
\end{table}

\subsubsection{Vision Data}
We use RGB images and object classes from the ILSVRC2012 validation set \cite{ILSVRC15}. We choose this dataset as it has 1000 object classes and a variety of images per object class, and we want our model to work on many different object classes and object instances. The ImageNet object classes such as ``violin,'' ``suit,'' 
and ``quill'' are usually nouns that occur frequently in textual data in correspondence with other verbs such as ``play,'' ``wear,'' ``write.''

\subsubsection{Language Data}
We extracted sentences from Wikipedia containing the ImageNet object classes and used spaCy \cite{spacy2} to parse the sentences and extract corresponding verb-object pairs. We originally sought out to extract verb-object pairs from the common-sense knowledge base ConceptNet \cite{speer2017conceptnet}, which ended up being too small and was missing many valid verb-object pairings. We then decided to use Wikipedia instead for its large text corpus. However, the resulting dataset was highly noisy with 20,198 verb-object pairs, containing many abstract verbs such as ``name,'' ``feature,'' ``use,'' or nouns in the wrong word sense such as ``suit'' in ``follow suit'' and ``file suit'' that were not relevant for the natural language object retrieval task we were interested in.
Therefore, we manually annotated the verb-object pairs to retain only pairs that contain concrete verbs paired with nouns in the correct sense. This resulted in a dataset with 655 verb-object pairs over 50 verbs and 216 object classes. Example verb-object pairs from our dataset are shown in Table \ref{tab:verb-obj}.

\begin{table}[]
\begin{center}
\caption{Example training data (command-image pairs)}
\begin{tabular}{lll}
\toprule
Verb-Object & Language Command & Image\\
\midrule
\multirow{2}{*}{contain -- cup} & \textit{Give me an item that can contain} & \includegraphics[width=0.15\linewidth]{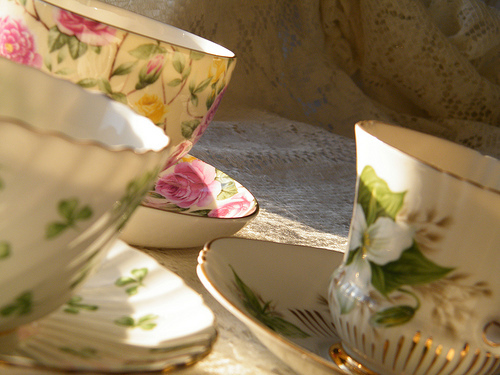}\\
& \textit{I need something to contain} & \includegraphics[width=0.15\linewidth, height=0.12\linewidth]{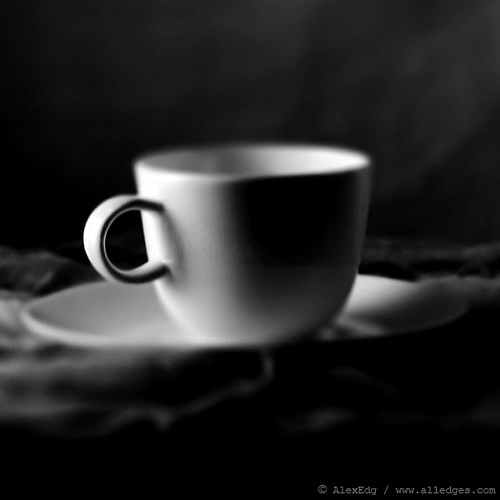}\\
\multirow{2}{*}{play -- drum} & \textit{Hand me something to play} & \includegraphics[width=0.15\linewidth]{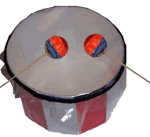}\\
& \textit{I want an object to play} & \includegraphics[width=0.15\linewidth]{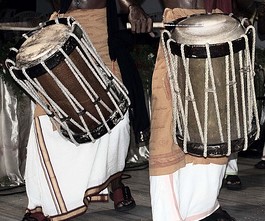}\\
\multirow{2}{*}{wear -- kimono} & \textit{An item to wear} & \includegraphics[width=0.15\linewidth, height=0.16\linewidth]{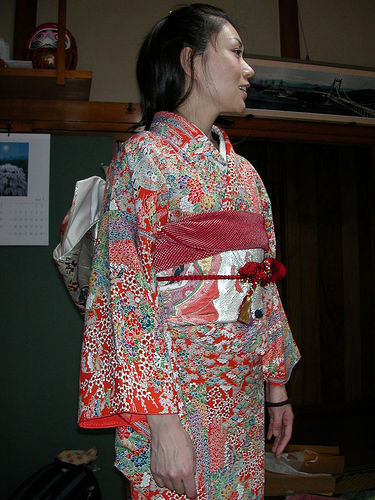}\\
& \textit{Give me something to wear} & \includegraphics[width=0.15\linewidth, height=0.13\linewidth]{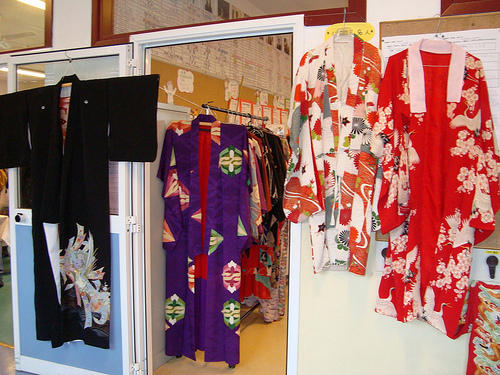}\\
\bottomrule
\label{tab:train-data}
\end{tabular}
\end{center}
\end{table}

\subsubsection{Training and Testing Data}
\label{sec:data}

We use 80\% of the 216 object classes and their corresponding verb-object pairs to generate our training data. The training data consist of natural language command-RGB image pairs. For each verb-object pair, language commands are generated from the pair using templates, such as $\texttt{<Hand me something to> <verb>}$, and then paired with different image instances of the object class. Examples of the training data are shown in Table \ref{tab:train-data}. Rather than using only the verbs and/or nouns from the verb-object pairs as language data for our model, we generate and give our model natural language sentences so that it can handle more realistic language commands, as most people do not ask for objects with one or two-word commands such as ``knife'' or ``knife cut.''

We hold out 20\% of the object classes for testing. Test examples consisting of language command-set of five images pairs are randomly generated from objects in the test set and their corresponding verb-object pairs. The evaluation task is to select the correct object from a set of five images given a natural language command.

\begin{figure*}[htbp]
\centering
\subfigure[Top-1 retrieval accuracies (\%)]{\includegraphics[width=0.47\textwidth]{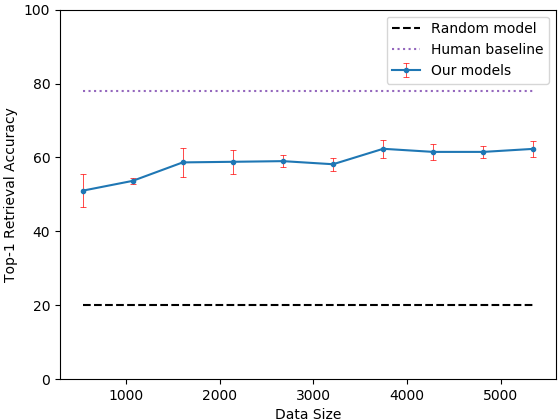}\label{fig:ret1}}\hspace{0.03\textwidth}
\subfigure[Top-2 retrieval accuracies (\%)]{\includegraphics[width=0.47\textwidth]{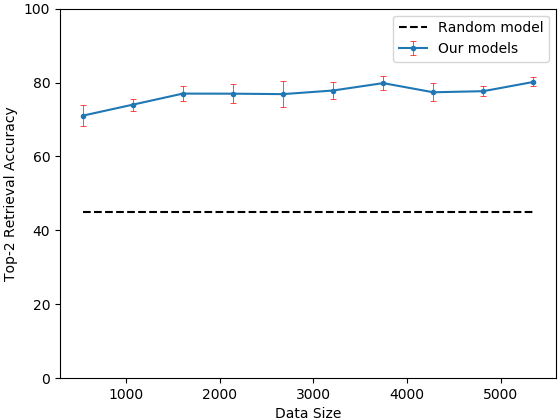}\label{fig:ret2}}
\caption{Average retrieval accuracies (with unseen object classes from the held-out object set) of models trained on different data sizes, represented by the solid lines, with vertical bars denoting standard errors. Models trained on larger data sizes usually perform better. All our models significantly outperform a random model, represented by the dashed lines. The dotted line represents the human object retrieval baseline.} \label{fig:ret}
\end{figure*}

\section{Experiments and Results}
The aim of our evaluation is to test our model's ability to accurately select objects based on natural language descriptions of their usage specified by verbs, given unseen object classes and unknown nouns in the language commands. Generalization to unseen object classes is much more difficult than just to unseen instances of known object classes, as different instances of the same object class such as two bottles would usually look more alike than instances from different object classes such as a bottle and a bowl, even if those object classes can be used for similar tasks such as ``contain."

The trained model is tested on natural language object retrieval tasks: retrieving the correct object from a set of five images of different objects, given a natural language command containing a verb that can only be paired with the correct object. The evaluation task is modeling a typical retrieval task in the wild where there are a few objects on a table and the robot has to pick the correct one. Retrieval examples consisting of a language command paired with a set of five images are randomly generated from objects in the test set and their corresponding verb-object pairs. We test our models on several different test sets and report average top-1 and top-2 retrieval accuracies. Top-1 accuracy means that the model's top choice is the correct answer, and top-2 accuracy means that the correct answer is among the model's top-2 choices.

\begin{table}[]
\begin{center}
\caption{Retrieval accuracies (\%) on unseen object classes in the held-out object set}
\begin{tabular}{lll}
\toprule
Model & Top-1 \textit{(Std. Error)} & Top-2 \textit{(Std. Error)}\\
\midrule
Random & 20.0 & 45.0\\
Data size 535 & 51.0 \textit{(4.50)} & 71.0 \textit{(2.84)}\\
Data size 1070 & 53.7 \textit{(0.86)} & 74.0 \textit{(1.60)}\\
Data size 1605 & 58.7 \textit{(3.85)} & 77.0 \textit{(2.13)}\\
Data size 2140 & 58.8 \textit{(3.31)} & 77.0 \textit{(2.50)}\\
Data size 2675 & 59.0 \textit{(1.68)} & 76.9 \textit{(3.55)}\\
Data size 3210 & 58.2 \textit{(1.70)} & 77.8 \textit{(2.30)}\\
Data size 3745 & \textbf{62.3 \textit{(2.48)}} & 79.8 \textit{(1.94)}\\
Data size 4280 & 61.5 \textit{(2.25)} & 77.4 \textit{(2.44)}\\
Data size 4815 & 61.5 \textit{(1.71)} & 77.7 \textit{(1.43)}\\
Data size 5350 & \textbf{62.3 \textit{(2.18)}} & \textbf{80.2 \textit{(1.23)}}\\
Human baseline & 78.0 \textit{(1.72)} & \\
\bottomrule
\label{tab:heldout}
\end{tabular}
\end{center}
\end{table}



\subsection{Held-out Object Set}
We first test our model on object sets held out from our dataset, which have a similar image distribution to that of the training set, as the images all come from the ILSVRC2012 validation set. We evaluate our model on 2 different test splits representing increasingly difficult scenarios. We describe the test splits and our model's performance in each case below. For both cases, the test objects are held-out, meaning our model has never seen any instances belonging to the test object classes during training.


\subsubsection{Unseen Object Classes}
\label{sec:unseen}

We first train and test our model on natural language commands containing only the verbs from the verb-object pairs, such as ``Hand me something to $\texttt{<verb>}$." This simplified setting where the test commands look like those in the training data, for example ``Give me something to contain,'' helps us look at how well the model has learned to generalize the concepts associated with the verbs, such as ``contain'' requires objects with convexity. It removes the additional challenges that might arise with seeing unseen nouns in the commands, such as the fact that the embeddings for these nouns would be untrained. However, this is still a challenging problem because the object classes in the test set have never been seen by the model before.


With 20\% of the objects and corresponding verb-object pairs in our dataset held out for testing, the training set contains 535 verb-object pairs. We trained separate models on increasing sizes of training data generated from the 535 verb-object pairs. Training data was augmented by generating different natural language commands containing the verbs, and pairing the commands with different images of the objects from the verb-object pairs in the training set. Examples of the training data are shown in Table \ref{tab:train-data}. The test set with 43 held-out objects and 120 corresponding verb-object pairs and retrieval examples are fixed for all models.

We tested each model trained on different data sizes 5 times and report their average top-1 and top-2 retrieval accuracies and standard errors in Table \ref{tab:heldout} and Figure \ref{fig:ret}. Model performance generally increases with larger training size. All our models significantly outperform a random model (which has 20\% top-1 and 45\% top-2 retrieval accuracy) and achieve accuracies in the 50\% -- 62\% range for top-1, and 70\% -- 80\% for top-2. Our best average retrieval accuracy is 62.3\% for top-1 and 80.2\% for top-2 with standard errors of 2.18\% and 1.23\%, respectively. Our models were able to generalize to unseen object classes.

Our models were able to select the correct object to satisfy the task specified by most verbs in our dataset such as ``contain," ``write," ``don," ``rotate," ``hit," etc. The objects paired with these verbs usually have similar visual appearances and attributes, for example a gown and a suit can both be paired with ``don" and are both made of fabric. However, our models performed imperfectly on more abstract verbs such as ``play" and ``protect" as it is less obvious which object attributes are required for the tasks specified by these verbs, and the objects that can satisfy the tasks come in a larger variety of visual appearances, for example a harp and a volleyball can both be paired with ``play."

\begin{figure*}[htbp]
\centering
\subfigure[Model success cases. The solid boxes denote the model's top choice for each task.]{\includegraphics[width=0.47\textwidth]{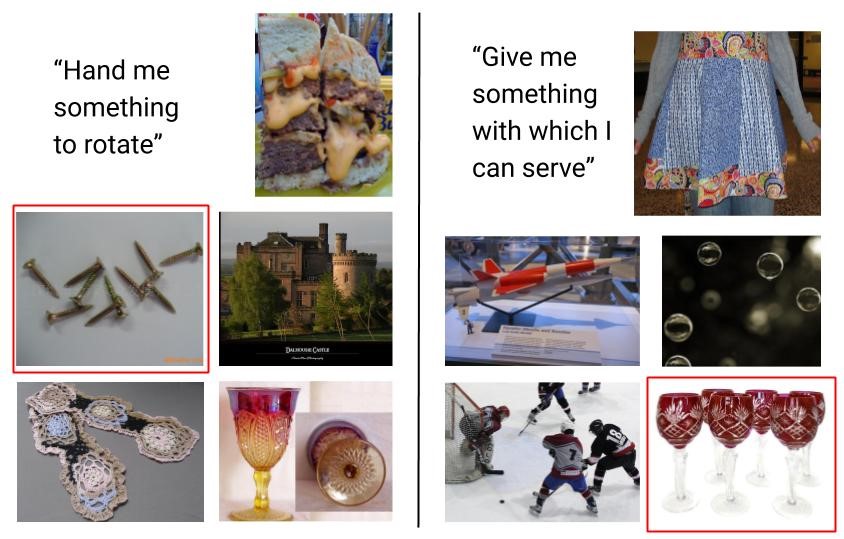}\label{fig:ex11}}\hspace{0.03\textwidth}
\subfigure[Model failure cases. The solid boxes denote the ground truth image that satisfies each command, and the dashed-line boxes denote the images the model actually chose.]{\includegraphics[width=0.47\textwidth]{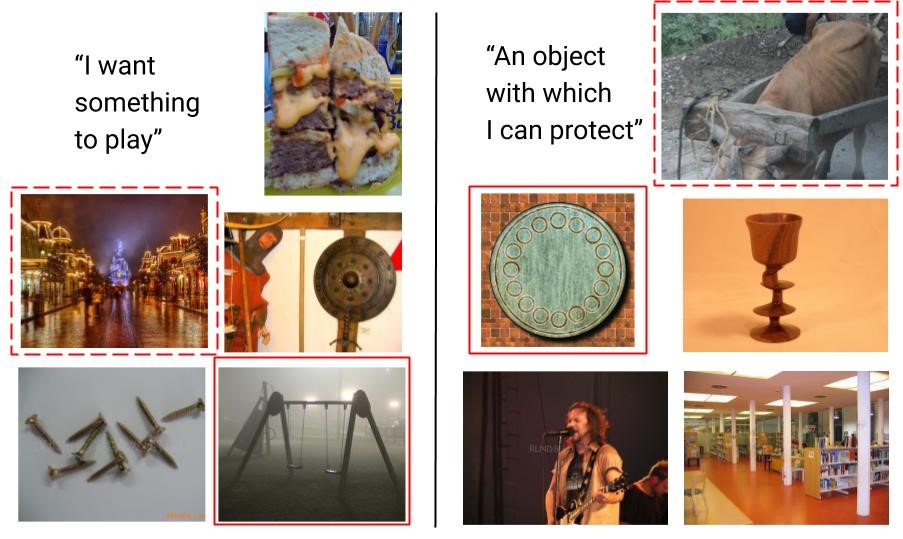}\label{fig:ex12}}
\subfigure[Model success cases. The solid boxes denote the model's top choice for each task.]{\includegraphics[width=0.47\textwidth]{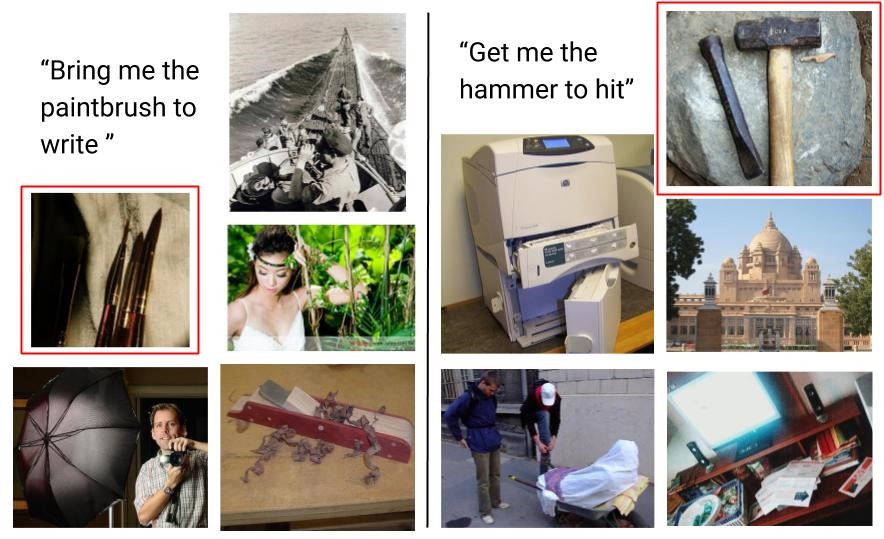}\label{fig:ex21}}\hspace{0.03\textwidth}
\subfigure[Model failure cases. The solid boxes denote the ground truth image that satisfies each command, and the dashed-line boxes denote the images the model actually chose.]{\includegraphics[width=0.47\textwidth]{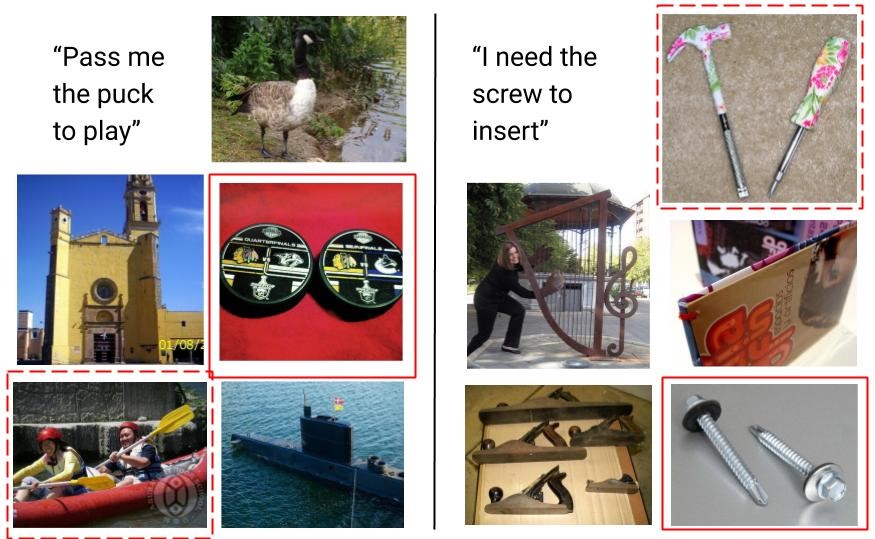}\label{fig:ex22}}
\caption{Example success and failure cases for our best models on object retrieval tasks with unseen object classes from the held-out object set, shown in (a) and (b), and with both unseen object classes \textit{and} unknown nouns, shown in (c) and (d). The models were given a natural language command and selected the object that they determine to best satisfy the command out of a set of five objects.} \label{fig:example1}
\end{figure*}

Example success and failure cases for our best model are shown in Figures \ref{fig:ex11} and \ref{fig:ex12}. Our model correctly selected images of screws and goblets to satisfy natural language commands ``Hand me something to rotate" and "Give me something with which I can serve," respectively. The model failed to select the swing given the command ``I want something to play," and did not select the shield in response to ``An object with which I can protect." However, the instance of a shield in the object retrieval task shown in Figure \ref{fig:ex12} does not look like a shield but more like a plate, and such object instance outliers can definitely throw the model off.

\subsubsection{Unseen Object Classes and Unknown Nouns}
Next, we train and test our model on natural language commands containing both verbs and objects, for example ``Give me the $\texttt{<object>}$ to $\texttt{<verb>}$." The model is tested on object retrieval tasks with both unseen object classes \textit{and} unknown nouns. Testing our model in this setting is necessary because when a deep net such as our model is dealing with an unknown word, it will map the word to a random, untrained embedding. That random embedding could completely throw off the model's understanding, for example the model might pick the image for whatever noun the random embedding happens to be closest to. We need to know how our model would behave with truly unknown words in the input to get a sense of how it would work in the real world.

An example task in this setting is the model getting the command ``Give me the dax to cut" with ``dax" being an unknown word to the model, while also being shown objects it has never seen before.
This task is more difficult than object retrieval with only unseen object classes, as the model has to figure out that unknown words such as ``dax" adds no information and avoid being affected by the noise added by the unknown words.

Other than the inclusion of nouns in the natural language commands, the setup for this experiment is the same as that with only unseen object classes, as described in Section \ref{sec:unseen}. Each model trained on different data sizes was tested 5 times. Our best average retrieval accuracy is 53.0\% for top-1 and 72.8\% for top-2 with standard errors of 1.33\% and 3.11\%, respectively. Our models were indeed negatively affected by the unknown words in the language commands. However, decline in performance is to be expected as this is a more difficult task. Furthermore, our models' performance still demonstrate at least some generalization to unseen object classes \textit{and} unknown nouns. A way to better handle unknown words and boost model performance in this setting would be to use pretrained word embeddings such as Word2vec \cite{mikolov2013efficient} or GloVe \cite{pennington2014glove} instead of random untrained embeddings.

Example success and failure cases for our best model are shown in Figures \ref{fig:ex21} and \ref{fig:ex22}. Our model was able to correctly select the paintbrush when asked to ``Bring me the paintbrush to write," and picked the hammer to satisfy the command ``Get me the hammer to hit." Unfortunately, the model incorrectly selected the canoe and hammer given the commands ``Pass me the puck to play," and ``I need the screw to insert," respectively. However, from the given image, canoeing does seem like a fun activity and maybe even something to ``play." In addition, the image representing the hammer also includes a screwdriver, an object that can be used to ``insert."

\subsection{Human Retrieval Baseline}

\begin{figure}[t]
\begin{center}
\includegraphics[width=1.0\linewidth]{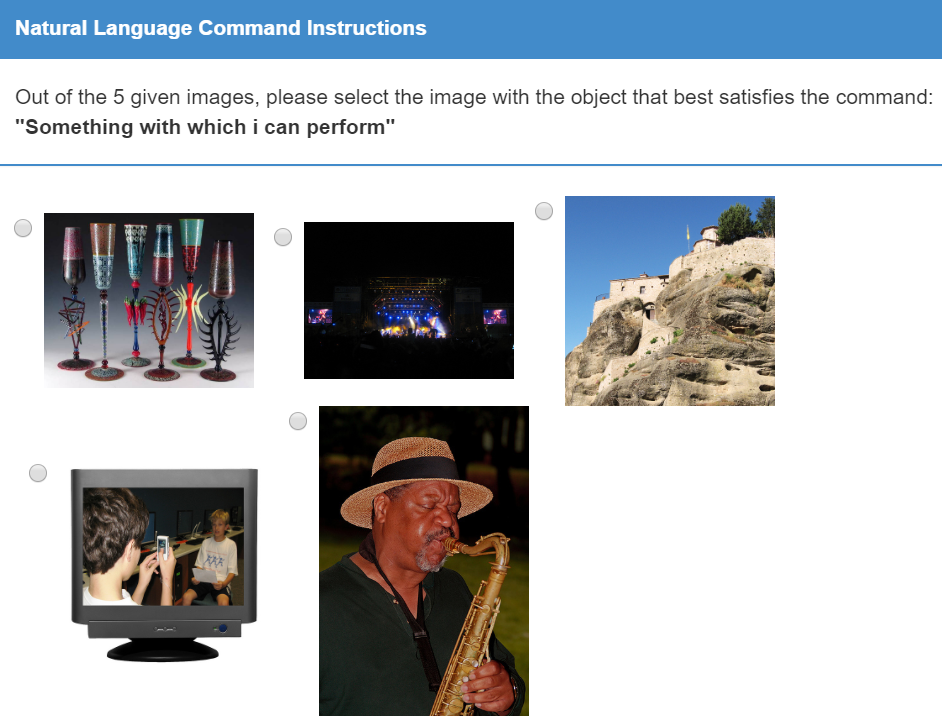}
\end{center}
\caption{Amazon Mechanical Turk interface and example task for human baseline experiment.}
\label{fig:amt}
\end{figure}

We also compare our models' performance to a human baseline for the retrieval task. Humans are experts in natural language understanding and object grounding. The experiment was done on Amazon Mechanical Turk (AMT). We showed AMT workers five images and one language command such as ``Give me something to $\texttt{<verb>}$," and asked them to select the image with the object that best satisfies the command. The AMT interface and an example task for the experiment is shown in Figure \ref{fig:amt}. We collected 5 answers for each of the 120 retrieval tasks. The average top-1 human retrieval accuracy is 78.0\% with standard error of 1.72\%, shown in Table \ref{tab:heldout} and Figure \ref{fig:ret1}. Even human users are not perfect at this task, as the given images of objects are not segmented and thus sometimes it can be confusing as to what object the image is supposed to be capturing, or the image only shows an partial/low-quality view of the object. In addition, the object usage being asked for in the language command can occasionally be unconventional such as using a ``spoon" to ``cut," and thus might not be obvious to the average AMT worker who is spending very little time on each task. Our models' performances are not as good as the human baseline but not far apart. Furthermore, the imperfect human performance proves how difficult of a task this is and how impressive our models' results are.

\subsection{YCB Object Set}


Finally, we run an evaluation to test whether the proposed model can perform natural language object retrieval on objects commonly seen and interacted with by real robots. For this evaluation, we test our best model on images of objects from the YCB Object and Model Set \cite{calli2015ycb}. The YCB object set is designed for benchmarking robotic manipulation and consists of objects of daily life with different shapes, sizes, textures, etc. We did not use the YCB object set as our training image set because it has a much smaller number of object classes and only 1 instance per object class in comparison to ImageNet's 1000 object classes and 50 images per class.

\begin{table}[]
\begin{center}
\caption{Verbs annotated with YCB objects}
\begin{tabular}{llll}
\toprule
construct & eat & open & serve\\
contain & grow & play & write\\
cut & hit & rotate & \\
\bottomrule
\label{tab:ycb}
\end{tabular}
\end{center}
\end{table}

\begin{table}[]
\begin{center}
\caption{Example annotated verb-object pairs for the YCB set}
\begin{tabular}{lll}
\toprule
construct -- power drill & eat -- apple & play -- tennis ball\\
contain -- chips can & grow -- pear & rotate -- adjustable wrench\\
contain -- windex bottle & hit -- spoon & serve -- bowl\\
cut -- scissors & open -- padlock & write -- large marker\\
\bottomrule
\label{tab:ycb-vo}
\end{tabular}
\end{center}
\end{table}

Of the 65 object classes with corresponding RGB images in the YCB dataset, we select 33 object classes to test our model on, excluding classes our model has seen during training and picking only one class in the case of identical objects of differing sizes such as ``S clamp," ``M clamp," ``L clamp," ``XL clamp." Each object class in the YCB dataset is represented by one object instance, with corresponding RGB images of the object instance from multiple camera angles. We represent each selected object class by a single front-facing image of the object, taken from the YCB dataset. From the 50 verbs our model was trained on, we select 11 verbs (shown in Table \ref{tab:ycb}) that are most compatible with the 33 YCB objects and annotated valid verb-object pairings among the selected objects and verbs, resulting in 64 verb-object pairs. Examples of the annotated verb-object pairs are shown in Table \ref{tab:ycb-vo}. 
Natural language commands containing only the verbs were generated from the verb-object pairs using templates, and retrieval examples consisting of sets of five images paired with language commands were randomly generated from the annotated verb-object pairs.

Our model, without being retrained on images from the YCB dataset, was tested 5 times and achieved average retrieval accuracies of 54.7\% and 71.9\% with standard errors of 1.99\% and 2.40\% for top-1 and top-2, respectively. Although these results are far from perfect, they still demonstrate generalization on a dataset with a different distribution from the model's training data. In addition, these results would enable the robot to significantly reduce its search space from all the candidate objects, and can employ strategies such as question asking to further disambiguate and retrieve the correct object.

Notably, our model correctly identified a fork, a spoon, and scissors as objects that can be used to cut, while only having seen knife-like object classes such as cleavers and hatchets paired with the verb ``cut" in its training data. In addition, our model selected a chips can and mustard bottle when asked for something to ``eat," which in retrospect are very reasonable pairings that we mistakenly left out of our verb-object pair annotations.

\subsection{Robot Demonstrations}

\begin{figure}[t]
\begin{center}
\includegraphics[width=1.0\linewidth]{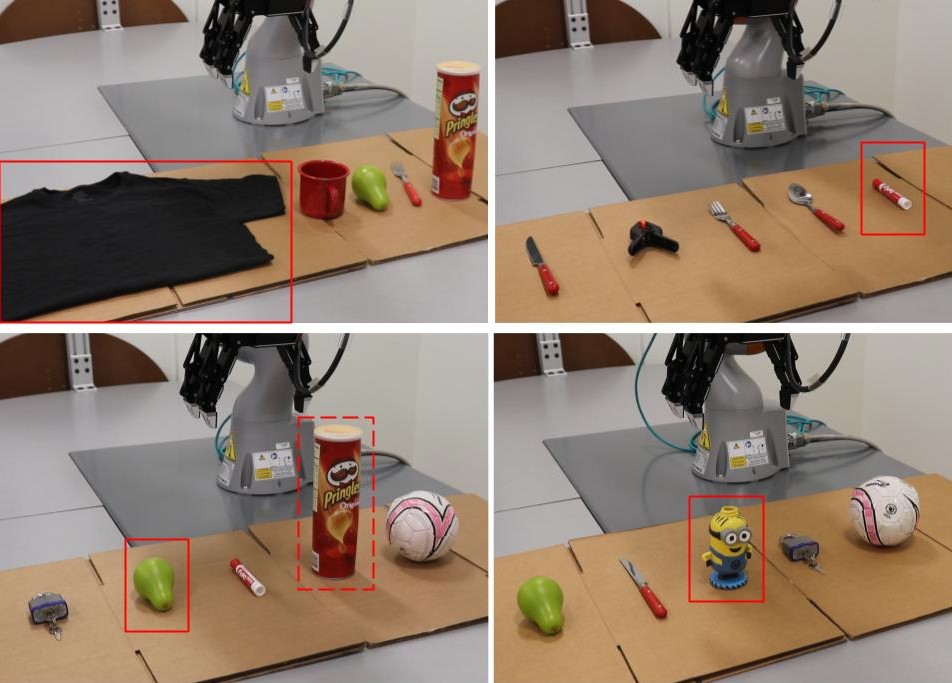}
\end{center}
   \caption{Natural language object retrieval tasks demonstrated on our robot. The given language commands are: ``Give me something to wear" (top left), ``Give me an item that can write" (top right), ``Hand me something to eat" (bottom left), and ``An object to contain" (bottom right). Solid boxes denote the robot's top choice for each task. The dashed-line box denotes the robot's top second choice. Our robot retrieved the correct object for each task.}
\label{fig:robot2}
\end{figure}

We implement our trained model on a KUKA LBR iiwa robot arm with a Robotiq 3-finger adaptive gripper.
We pass a natural language command into our model along with manually segmented RGB images of objects in the scene, captured by an Intel RealSense camera.
The robot then grasps
the observed object with the highest cosine similarity in embedding space with the language command. We use object classes that our model has not seen during training for the demonstrations.

We tested our robot on four object retrieval tasks. Images capturing the tasks are shown in Figure \ref{fig:robot2}. The robot correctly selected the T-shirt for the task of ``Give me something to wear." When asked to ``Give me an item that can write," it was able to pick out the marker from other distracting objects that are also partly red and have slim bodies. Next, it accurately identified the pear and chips can as the top two items that would satisfy ``Hand me something to eat." This is the only test case with more than one possible correct answer. Finally, when asked for ``An object to contain," the robot selected the empty Minion-shaped bottle. Video recordings of the robot demonstrations can be found online\footnote{\url{https://youtu.be/WMAdGhMmXEQ}}.

We mostly use YCB objects for the demonstrations with the exception of the Minion-shaped bottle in the last case, which was to test our model on an odd-looking object. While most of these are common objects we see in our daily life, not all of them belong to the COCO dataset of common objects in context \cite{lin2014microsoft}. The objects that are not part of the 91 object types in COCO are the T-shirt, marker, pear, clamp, lock, and of course Minion bottle. As it was trained on the COCO dataset, Mask R-CNN \cite{he2017mask}, the state-of-the-art method for object segmentation and classification,
was unable to correctly segment and classify these six objects.
With such classification results, relying on accurate classification of objects and querying of an external knowledge base for valid verb-object pairs to select the object that satisfies the language command does not work in these cases. In contrast, our model was able to select the correct object based on the command without needing to explicitly classify the candidate objects or having seen the object classes.

\section{Conclusion}
Understanding open-ended natural language commands is a challenging but important problem. We address a sliver of the problem by focusing on object retrieval based on descriptions of the object's usage. We propose an object retrieval model that learns from contextual information from both language and vision to generalize to unseen object classes and unknown nouns. Given natural language commands, our model correctly selects objects out of sets of five candidates, and achieves an average accuracy of 62.3\% on a held-out set of unseen ImageNet object classes 
and 53.0\% on unseen object classes \textit{and} unknown nouns. Our model also achieves an accuracy of 54.7\% on unseen YCB object classes. We demonstrate our model on a KUKA LBR iiwa robot arm, enabling the robot to retrieve objects based on natural language descriptions of their usage. Along with our model, we also present a newly created dataset of 655 verb-object pairs denoting object usage over 50 verbs and 216 object classes, as well as the methods used to create this dataset. To the best of our knowledge, this is the first dataset built to perform this task, and could potentially be used for a range of object retrieval tasks.

Our model currently allows us to reduce the problem of task based object retrieval  to an attribute classification problem. However, a much richer model would perform explicit inference to determine the desired object from oblique natural language. Incorporating dialogue into this framework to perform inference can be a way to incorporate human preference more directly and provide a more intuitive interface. 

\section{Acknowledgments}
The authors would like to thank Prof. James Tompkin for advice on selecting the image dataset and encoder, and Eric Rosen for help with video editing. This work is supported by the National Science Foundation under award numbers IIS-1652561 and IIS-1717569, NASA under award number NNX16AR61G, and with support from the Hyundai NGV under the Hyundai-Brown Idea Incubation award and the Alfred P. Sloan Foundation.


\bibliographystyle{plainnat}
\bibliography{ref}

\end{document}